%% file: main.tex
\documentclass{article}
\usepackage{spconf,amsmath,amssymb,graphicx,booktabs,xcolor}
\usepackage[T1]{fontenc}
\usepackage{microtype,placeins,xurl}
\usepackage[hidelinks]{hyperref}
\usepackage{orcidlink}
\newcommand{\E}{\mathbb{E}}
\newcommand{\ind}{\mathbf{1}}

\newcommand{\aulcresult}[3]{#1\(\pm\)#2\,(#3)}
\makeatletter
\g@addto@macro\small{%
  \setlength{\abovedisplayskip}{4pt plus 1pt minus 2pt}%
  \setlength{\belowdisplayskip}{4pt plus 1pt minus 2pt}%
  \setlength{\abovedisplayshortskip}{1pt plus 1pt}%
  \setlength{\belowdisplayshortskip}{2pt plus 1pt minus 1pt}}
\renewcommand\section{\@startsection{section}{1}{\z@}{-\baselineskip}{\baselineskip}{\normalfont\bfseries}}
\renewcommand\subsection{\@startsection{subsection}{2}{\z@}{-8pt plus -2pt minus -1pt}{4pt plus 1pt}{\normalfont\bfseries}}
\makeatother

\title{Should Pseudo-Labels Share Mini-Batches with Labeled Examples?\\Semi-Supervised Continual Active Learning with Limited Training Cost}
\name{Shiqi Zhang\,\orcidlink{0009-0008-6012-8088}, Tuomas Virtanen\,\orcidlink{0000-0002-4604-9729}}
\address{Faculty of Information Technology and Communication Sciences\\
  Tampere University, Tampere, Finland}

\begin{document}
\ninept
\def\baselinestretch{0.927}\normalsize
\maketitle

\begin{abstract}
    Active learning (AL) reduces annotation by querying informative examples.
    Continual active learning (CAL) also cuts the training cost, the total number of gradient steps, by training one model across AL rounds with a fixed number of mini-batch steps per round.
    Pseudo-labels add supervision without annotation, but they increase training cost.
    In CAL, where every round trains on a fixed number of mini-batches, whether and how pseudo-labeled examples should share these mini-batches with labeled examples is therefore open.
    We provide the first formalization of this problem as two decisions after each query, namely (i) which unlabeled examples should enter training with a pseudo-label and (ii) what share of each mini-batch they should take.
    Our method, semi-supervised continual active learning (SSCAL), works with any acquisition function.
    It admits pseudo-labels based on distances from unlabeled examples to labeled neighbors, caps their share, and reweights the loss on labeled examples to keep their expected gradient contribution unchanged.
    On frozen embeddings, its pseudo-labeler updates incrementally without extra model forward passes.
    Across two audio datasets, two frozen encoders, six acquisition functions and ten seeds, SSCAL ranks first on average among four protocols with equal training steps, answering the question affirmatively.
\end{abstract}


\section{Introduction}
\label{sec:intro}
Active learning (AL) selects informative examples to label and thereby reduces the annotation that model training needs.
In audio, for instance, AL has studied clustering with label propagation~\cite{zhao2017audio}, confidence sampling~\cite{wang2019audio}, and disagreement with diversity~\cite{zhang2025audio}.
These advances target annotation efficiency.
As models grow larger and deeper, the training cost, the total number of gradient steps spent over all AL rounds, becomes a concern of its own.
In AL this cost also delays annotation.
When the queries depend on the model, the annotator needs to wait for training in every round before labeling the next queries.
Conventional AL protocols retrain for a fixed number of epochs in each AL round, so their training steps are proportional to the labeled set, which grows every round, and the cumulative cost over \(T\) rounds grows quadratically in~\(T\).
To avoid this quadratic cost, continual active learning (CAL)~\cite{das2023cal} continues training the same model across rounds instead of reinitializing it (model continuation) and keeps training on earlier labels alongside the new ones (replay).
The simplest replay in CAL combines new labeled examples with uniformly sampled earlier ones in every mini-batch~\cite{das2023cal}.
Uniform replay, which samples every labeled example with equal probability, matches or outperforms costlier sampling and distillation methods in fixed-step continual learning~\cite{prabhu2023budgetcl}.

Semi-supervised learning has long combined the supervised and pseudo-label losses within each mini-batch, with a scheduled weight on the pseudo-label loss~\cite{lee2013pseudo} or a fixed ratio of unlabeled to labeled examples~\cite{sohn2020fixmatch}.
Its pseudo-labels come from the predictions of the model being trained~\cite{lee2013pseudo,sohn2020fixmatch} or from labeled neighbors in a feature space, by label propagation over a graph of the examples~\cite{zhu2002propagation,zhou2004consistency,iscen2019propagation} or by nearest-neighbor assignment from labeled support examples~\cite{assran2021paws}.
Semi-supervised AL borrows this practice to also train on unlabeled examples, through consistency regularization~\cite{gao2020consistency}, pseudo-labels~\cite{simeoni2020rethinking,yin2025synergy} or self-training~\cite{shishkin2021audio}.
In both lines of work the labeled set is fixed during training and the training cost is usually unconstrained, so pseudo-labeled examples add computation on top of the labeled ones and crowd out none of them.

Inspired by both lines of work, we study whether pseudo-labels can also help CAL when every round is fixed at \(K\) gradient steps on stochastic mini-batches of \(m\) examples, which keeps the cumulative cost linear in~\(T\). We call this setting fixed-step CAL.
Under this constraint, pseudo-labels replace labeled supervision instead of adding to it.
A round samples exactly \(Km\) examples into its mini-batches, so each pseudo-labeled example among them replaces a labeled one.
Labeled and pseudo-labeled examples thus share every fixed-size mini-batch.
The displacement worsens once the labeled set grows past \(Km\), after which a round samples fewer examples than those are labeled, and pseudo-labels, which can be incorrect, crowd out supervision that is already insufficient.
Whether pseudo-labels are worth admitting, and what share of each mini-batch they may take, thus depends on how accurate and numerous the admitted pseudo-labels are and on how much labeled supervision they displace.
Without a training constraint this trade-off does not arise, so the additive gains of semi-supervised learning do not settle it.
We therefore examine this mini-batch sharing in AL at an unchanged annotation budget, where labels are scarce, arrive incrementally, and change every round which pseudo-labels the labeled examples support.

We formalize this problem as two decisions taken after each query and define the corresponding protocol.
First, pseudo-label admission decides which unlabeled examples enter training with a pseudo-label.
Second, mini-batch allocation decides how the mini-batches are shared by setting the sampling distribution over labeled and pseudo-labeled examples.
We propose semi-supervised continual active learning (SSCAL), a method that works with any acquisition function.
It admits a pseudo-label when an unlabeled example lies closer to two labeled positives of a class than to the nearest labeled negative of that class, caps the sampling share of pseudo-labeled examples, and reweights the loss on labeled examples to keep their expected gradient contribution equal to that of uniform replay.
Its pseudo-labeler runs incrementally on frozen embeddings without extra model forward passes.
Among four protocols with equal training steps, \mbox{SSCAL} attains the highest area under the learning curve in 202 of 240 paired runs and a mean rank of 1.18.

Our contributions are (i) to our knowledge, the first study to bring pseudo-labels into fixed-step CAL, formalized as pseudo-label admission and mini-batch allocation with a corresponding protocol;
(ii) SSCAL, an acquisition-function-agnostic method for both decisions;
and (iii) experimental validation that answers the mini-batch sharing question affirmatively.

\section{Problem and protocol}
\label{sec:problem}
Pool-based AL starts with a pool \(\mathcal{D}=\{x_1,\ldots,x_N\}\) of \(N\) examples \(x_i\), a fixed label space of \(C\) classes, and an annotator who returns the label \(y_i\) of any queried example.
All examples are available for training and for computing queries, whereas labels are available only for queried examples.
Pseudo-labels may use the acquired labels, the examples themselves and the current model.
AL round~\(t\) starts from the labeled set \(L_{t-1}\), where \(L_0\) is a random initial labeled set of \(n_0\) examples, acquires labels for a query set \(Q_t\) of \(b\) unlabeled examples chosen by the acquisition function, and sets \(L_t=L_{t-1}\dot\cup Q_t\) and \(U_t=\mathcal{D}\setminus L_t\).
We write \(H_t=L_{t-1}\) for the earlier labels, so \(L_t=H_t\dot\cup Q_t\).
The model \(f_\theta\) is trained to output class logits, and its parameters and optimizer state carry over rounds.
Each round performs \(K\) gradient steps, each on a stochastic mini-batch of \(m\) examples.
Since \(K\) and \(m\) are fixed, each round samples \(Km\) examples.
The acquisition function is an external choice held fixed over all rounds, and its queries may depend on the current model.

After each query, a policy pair \(\Pi=(\pi_p,\pi_s)\) makes the two decisions.
The admission policy \(\pi_p\) selects a set \(P_t\subseteq U_t\) of pseudo-labeled examples and assigns each a pseudo-label target \(\widehat z_{t,i}\).
The allocation policy \(\pi_s\) sets the distribution from which mini-batches are sampled.
Training uses the union \(A_t=H_t\dot\cup Q_t\dot\cup P_t=L_t\dot\cup P_t\) of labeled and pseudo-labeled examples, each with a target \(z_{t,i}\) that is the acquired label \(y_i\) on \(L_t\) and the pseudo-label \(\widehat z_{t,i}\) on \(P_t\).
Pseudo-labeled examples consume no annotation budget and stay in~\(U_t\), so they remain queryable.
To simplify notation, the step index within~a~round is omitted.

The policy \(\pi_s\) sets a distribution \(\nu_t\) over \(A_t\), and the \(m\) examples of a mini-batch are sampled i.i.d.\ (with replacement) from \(\nu_t\), so an example may appear in one mini-batch more than once.
Example~\(i\) therefore appears \(m\nu_{t,i}\) times in a mini-batch in expectation, that is, it occupies an expected fraction \(\nu_{t,i}\) of every mini-batch, which we call its sampling share.
The sampling share \(\alpha_t^G\) of a group \(G\in\{H,Q,P\}\) is the expected fraction of a mini-batch that its members occupy together, and the three group shares sum to one because \(\nu_t\) is a distribution over the disjoint union \(A_t\) of the three groups,
\begin{equation}
    \alpha_t^G=\sum_{i\in G_t}\nu_{t,i},\qquad
    \alpha_t^H+\alpha_t^Q+\alpha_t^P=1.
    \label{eq:slots}
\end{equation}
Any share given to pseudo-labeled examples is thus taken from labeled examples, which is how the two share every mini-batch.

Admission, targets and sampling shares jointly determine the expected gradient of one step.
Let \(\ell_i=\ell(\theta;x_i,z_{t,i})\) be the loss of example~\(i\) on its target, and let \(w_{t,i}\) be the loss weight of example~\(i\), which \(\pi_s\) may set to compensate for non-uniform sampling (\(w_{t,i}=1\) unless stated).
For the weighted mini-batch loss \(\widehat R_t=\frac{1}{m}\sum_{j=1}^{m}w_{t,i_j}\ell_{i_j}\) over the sampled examples \(i_1,\ldots,i_m\),
\begin{equation}
    \E[\nabla\widehat R_t]=
    \sum_{i\in L_t}\nu_{t,i}w_{t,i}\nabla\ell_i+
    \sum_{i\in P_t}\nu_{t,i}w_{t,i}\nabla\ell_i,
    \label{eq:gradient}
\end{equation}
where the first sum is the contribution of labeled examples and the second that of pseudo-labeled examples.

Since the number of AL rounds available in practice is unknown in advance, the objective is the normalized area under the learning curve (AULC) over all \(T\) rounds of a run, which averages the task metric over all labeled-set sizes,
\begin{equation}
    J(\Pi)=\E_\Pi\!\left[
        \frac{\sum_{t=2}^{T}(n_t-n_{t-1})(M_t+M_{t-1})/2}{n_T-n_1}
        \right],
    \label{eq:aulc}
\end{equation}
where \(n_t=|L_t|\), \(M_t\) is the task metric after round~\(t\) on test data held out from the acquisition function and the policies, and the expectation is over the randomness of a run.
With label counts, \(K\) and \(m\) fixed, policies are compared at equal annotation and training steps.

\section{Method}
\label{sec:method}
\input{tables/t1_main}
\input{tables/t2_strategy}
\begin{figure*}[t]
    \centering
    \includegraphics[width=\textwidth]{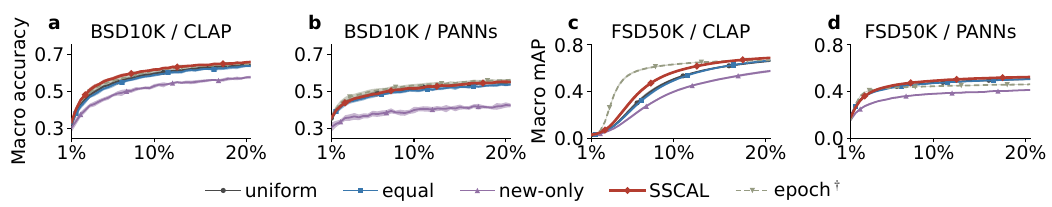}
    \caption{Aggregated test learning curves for the four settings against the labeled fraction of the training set. Each seed first averages six acquisition functions. Lines and shading show the mean and one standard deviation of these curves across ten seeds. The epoch\(^\dagger\) protocol runs \(10.5\times\) as many training steps and is shown for reference only.}
    \label{fig:curves}
\end{figure*}
\subsection{Neighbor-margin admission}
Independently of acquisition scores, SSCAL admits pseudo-labels by propagating acquired labels to nearby unlabeled examples in a frozen embedding space, a nearest-neighbor form of label propagation in semi-supervised learning~\cite{zhu2002propagation,zhou2004consistency,iscen2019propagation}.
Pretrained embeddings tend to place examples of the same class close together, so an unlabeled example likely shares the class of labeled examples that lie much closer to it than all other labeled examples.
Labeled neighbors are preferred to the predictions of the model being trained~\cite{lee2013pseudo,sohn2020fixmatch} because acquired labels are reliable and grow every round, whereas the model is trained for only \(K\) steps per round and its errors would feed back into its own training~\cite{arazo2020confirmation}.
On \(\ell_2\)-normalized embeddings, a labeled example is a positive of class~\(c\) if its acquired label contains \(c\) and a negative of class~\(c\) otherwise.
For an unlabeled candidate~\(i\), let \(\Delta^{+,2}_{i,c}\) and \(\Delta^-_{i,c}\) be the squared Euclidean distances from \(i\) to its second-nearest positive and to its nearest negative of class~\(c\).
SSCAL admits class~\(c\) as a pseudo-label of candidate~\(i\) when the second-nearest positive is closer than the nearest negative by a margin set by a threshold \(0<\rho\leq1\),
\begin{equation}
    \begin{gathered}
        r_{i,c}=\Delta^{+,2}_{i,c}/\Delta^-_{i,c},\qquad
        a_{i,c}=\ind\{r_{i,c}<\rho^2\},\\
        P_t=\{i\in U_t:\textstyle\sum_c a_{i,c}>0\},
    \end{gathered}
    \label{eq:admit}
\end{equation}
where \(r_{i,c}\) is the distance ratio and \(a_{i,c}\) the admission indicator.
Only candidates with at least one admitted class enter \(P_t\) and thus training, since the others lack the labeled support for a reliable pseudo-label.
Requiring two positives means that one mislabeled or atypical neighbor cannot admit a pseudo-label on its own, and the nearest negative bounds how far the positives may lie.
Class~\(c\) is not admitted (\(a_{i,c}=0\)) when fewer than two positives or no negative of class~\(c\) has been labeled, or when the nearest negative is at distance zero.
For single-label tasks, at most one class is admitted per candidate, since its second-nearest positive must be closer than the nearest labeled example of every other class.
The admitted class is that of the nearest labeled neighbor and is the cross-entropy target \(\widehat z_{t,i}\).
For multi-label tasks, the pseudo-label loss \(\ell_i=C^{-1}\sum_c a_{i,c}\operatorname{BCE}(f_\theta(x_i)_c,1)\) supervises only the admitted classes, each with target one, as the other classes remain unknown.
Labeled examples are trained with cross-entropy in single-label tasks and with binary cross-entropy averaged over all \(C\) classes in multi-label tasks.

Labeled negatives near a candidate tighten admission and thereby bound pseudo-label errors.
For class~\(c\), suppose every candidate that does not belong to \(c\) is at Euclidean distance at least \(s>0\) from every positive of \(c\) and within \(h>0\) of some negative of~\(c\).
Then \(\Delta^{+,2}_{i,c}\geq s^2\) and \(\Delta^-_{i,c}\leq h^2\), so \(\rho h\leq s\) makes the strict inequality in~\eqref{eq:admit} impossible and no false positive is admitted for~\(c\).
Because admission uses only labeled examples as references, a newly acquired negative can also revoke an earlier admission.

\subsection{Mini-batch allocation}
In AL the labeled examples are few and sparse relative to the size of the pool, so the pseudo-labels propagated from them are of limited reliability.
To keep most of every mini-batch for the more reliable acquired labels, SSCAL caps the sampling share of pseudo-labeled examples at a hyperparameter \(\beta\in[0,1)\), reserving at least \(1-\beta\) for labeled examples, and samples uniformly within each group of \(A_t\),
\begin{equation}
    \begin{gathered}
        \alpha_t^P=\min\!\left\{\beta,\frac{|P_t|}{|L_t|+|P_t|}\right\},\\
        \nu_{t,i}=\begin{cases}(1-\alpha_t^P)/|L_t|, & i\in L_t, \\
                     \alpha_t^P/|P_t|,     & i\in P_t,\end{cases}
    \end{gathered}
    \label{eq:sampling}
\end{equation}
where the minimum of the two keeps the share of a small \(P_t\) at no more than its proportion among all training examples.
Since labeled examples are now sampled with total probability \(1-\alpha_t^P\) instead of one, their losses are weighted by \(w_{t,i}=1/(1-\alpha_t^P)\).
Pseudo-labeled examples have no reference contribution to restore and keep unit weight.
Admission, targets and shares stay fixed within a round.

At any parameter value, this weighting preserves the expected gradient contribution of the labeled examples exactly as under uniform replay,
\begin{equation}
    \E[\nabla\widehat R_t]=
    \frac{1}{|L_t|}\sum_{i\in L_t}\nabla\ell_i+
    \frac{\alpha_t^P}{|P_t|}\sum_{i\in P_t}\nabla\ell_i,
    \label{eq:preserve}
\end{equation}
whose first term is the expected gradient of uniform replay over \(L_t\) and whose second term adds the pseudo-label gradient with total weight \(\alpha_t^P\).
Labeled examples are still sampled \(1-\alpha_t^P\) times as often as under uniform replay, so the \(K\) steps can follow a different trajectory.
This is the displaced supervision studied here.
The cap \(\beta\) and weight \(w_{t,i}\) play the roles of the unlabeled ratio and loss weight in additive semi-supervised learning~\cite{lee2013pseudo,sohn2020fixmatch}, and here jointly bound the displaced labeled supervision and restore its expected gradient.
With empty \(P_t\) or \(\beta=0\), SSCAL reduces to uniform replay.

\subsection{Incremental cost}
The frozen-embedding setting lets the pseudo-labeler update incrementally as labels are acquired.
For every candidate and class it stores three distances, to the two nearest positives and the nearest negative, and each round computes distances only to the \(b\) newly labeled examples.
A round thus costs \(O(Nb(d+C))\) for embedding dimension \(d\), independent of the size of the labeled set.
It needs no extra forward pass of \(f_\theta\) and leaves the \(K\) steps per round untouched.
Table~\ref{tab:diagnostics} reports the measured costs.

\section{Experiments}
\label{sec:experiments}
\subsection{Setup}
BSD10K (23 classes, macro accuracy)~\cite{anastasopoulou2024bsd,anastasopoulou2025bsd} and FSD50K (200 classes, macro mean average precision, mAP)~\cite{fonseca2022fsd50k} represent single-label and multi-label audio classification.
Fixed train, validation and test splits contain 8,767, 1,098 and 1,091 examples for BSD10K (a stratified 8:1:1 split) and 36,785, 4,169 and 10,230 for FSD50K (the official split excluding 13 clips longer than 30\,s).
The annotator is simulated by the dataset labels, revealed as examples are queried.
Frozen embeddings come from contrastively pretrained CLAP (Transformer, 512 dimensions)~\cite{wu2023clap} and supervised-pretrained PANNs CNN14 (CNN, 2,048 dimensions)~\cite{kong2020panns}, and a standard GELU MLP head \(d\to256\to C\) shared by all protocols is trained with Adam at learning rate \(5\times10^{-3}\) and zero weight decay.

A run consists of \(T=40\) AL rounds with \(K=20\) steps per round and \(b=m=n_0=44\) for BSD10K or 184 for FSD50K, that is, 0.5\% of the respective training set.
These values keep annotation sparse and training short.
The labeled set grows from 0.5\% to 20.5\% of the training set, and the \(Km=20b\) examples sampled per round cover it ten times in the first round and under half in the last.
Each of ten paired seeds fixes the same random initial labeled set \(L_0\) and model initialization for all protocols and acquisition functions.
As a reference from the conventional AL setting, \emph{epoch} makes \(E=10\) passes over \(L_t\) per round, \(10.5\times\) as many steps over a run.
Before the first AL round, all protocols train on \(L_0\), fixed-step protocols for \(K\) steps (an identical paired warm-up, so AULC in~\eqref{eq:aulc} starts at the first AL round) and epoch for \(E\) passes.
The acquisition functions are random, entropy, margin~\cite{settles2009survey}, Core-set~\cite{sener2018coreset}, TypiClust~\cite{hacohen2022typiclust} and BADGE~\cite{ash2020badge}, spanning uncertainty, representativeness and hybrid methods.
For multi-label tasks, entropy sums binary entropies, margin averages the five smallest distances to probability 0.5, and BADGE uses BCE gradient embeddings.
Entropy, margin and BADGE query with the current model, so protocols under the same seed and acquisition function may acquire different labels and the comparison covers the complete loop of querying, pseudo-labeling and training.

The baselines are \emph{uniform}, which samples \(L_t\) uniformly, \emph{equal}, which gives sampling share \(1/2\) each to \(H_t\) and \(Q_t\), and \emph{new-only}, which samples only \(Q_t\).
All use the same \(K\) steps of \(m\) examples per round as SSCAL.
The default \((\rho,\beta)=(1.0,0.4)\) maximizes mean validation AULC over all settings on the grid \(\rho\in\{0.6,0.8,1.0\}\), \(\beta\in\{0.1,0.2,0.4\}\) (Table~\ref{tab:ablation}).
The main comparison contains 1,200 runs (3,120 with the full grid).

\subsection{Performance across settings and acquisition functions}
SSCAL significantly improves over uniform in three of the four dataset/encoder settings (Table~\ref{tab:main}).
AULC rises by 2.00, 4.55 and 1.46 points for BSD10K/CLAP, FSD50K/CLAP and FSD50K/PANNs (paired two-sided Wilcoxon, \(p\leq5\times10^{-10}\)), while BSD10K/PANNs gains only 0.17 points (\(p=0.42\)).
SSCAL achieves the best mean rank with each acquisition function (Table~\ref{tab:strategy}), ten-seed means exceed uniform in 21 of the 24 dataset/encoder/acquisition configurations, and entropy and margin gain most, from 45.28 to 48.68 and from 46.20 to 49.64 AULC.

Gains emerge early and persist across rounds (Fig.~\ref{fig:curves}).
The equal and new-only baselines, which change only the sampling shares among labeled examples, reduce mean AULC across the four settings by 0.27--1.03 and 6.59--11.30 points, so the benefit comes from sharing the mini-batches with pseudo-labeled examples.
Interestingly, the epoch reference even falls below uniform on FSD50K/PANNs, so extra steps over the full labeled set do not guarantee gains even without the fixed-step constraint.

\subsection{Admission and sampling sensitivity}
\input{tables/t3_ablation}
To assess hyperparameter sensitivity we run all nine admission and sampling parameter combinations.
Every one achieves the best mean rank (Table~\ref{tab:ablation}), with mean ranks of 1.08--1.38 and aggregate AULC gains of 0.12--2.04 points.
Increasing \(\rho\) improves aggregate gains (\(\rho=1\) still requires both positives strictly closer than the nearest negative), and a larger \(\beta\) helps at \(\rho=1\) but not consistently under strict admission.

Counterintuitively, relaxing admission lowers pseudo-label precision while improving AULC in all four settings.
At \(\beta=0.4\), raising \(\rho\) from 0.6 to 1.0 lowers precision from 97.1/91.0/99.9/99.3\% to 83.2/75.4/95.2/87.2\% in the order of Table~\ref{tab:main}, while AULC rises from 57.97/50.69/44.75/45.64 to 59.79/50.86/48.95/47.15.
Pseudo-label precision alone thus does not predict the task benefit.
At the same cap \(\beta=0.4\), the mean sampling share \(\alpha_t^P\) rises from 0.19--0.38 to 0.40 and broader admission brings more unlabeled inputs into training, so this expanded coverage and larger sampling share may explain the gains.

\subsection{Diagnostics, cost and scope}
\input{tables/t4_diagnostics}
At the default parameters, pseudo-label quality depends on how well the representation separates the classes (Table~\ref{tab:diagnostics}).
BSD10K/PANNs has the lowest precision (75.4\%) and a neutral performance change.
Precision is computed offline by replaying each configuration's own query trajectory, so unacquired labels never enter the protocol.

Over initialization and 40 AL rounds, SSCAL and uniform run 820 steps and epoch runs 8,610.
Relative to uniform, SSCAL adds 0.25--1.36\,s of training-side wall time and epoch adds 7.43--15.95\,s, so the extra time behind SSCAL's gains stays far below that of the conventional epoch protocol.

\FloatBarrier
\section{Conclusion and future direction}
\label{sec:conclusion}
We formalize pseudo-label admission and mini-batch allocation in fixed-step CAL, define its protocol, and answer the sharing question affirmatively.
The acquisition-function-agnostic SSCAL method improves 21 of 24 audio configurations and ranks best on average under all six acquisition functions.
Future work could extend SSCAL to end-to-end training, refreshing distances as the embeddings change.

\clearpage

\section{Compliance with Ethical Standards}
\label{sec:ethics}
This study is limited to computational analysis of the existing public BSD10K~\cite{anastasopoulou2024bsd} and FSD50K~\cite{fonseca2022fsd50k} datasets.
No new experiments involving human or animal subjects were conducted.

\section{Acknowledgements}
\label{sec:acknowledgements}
This work received funding from the European Union's Horizon Europe programme under the Marie Sk{\l}odowska-Curie grant agreement No.~101116715 (BioacousticAI).

\bibliographystyle{IEEEtran}
\bibliography{refs}
\end{document}

%% file: tables/t1_main.tex
\begin{table*}[t]
    \centering
    \small
    \caption{AULC (\(\%\), higher is better), mean\(\pm\)s.d. over ten seeds, each averaging six acquisition functions. Parentheses give mean ranks among four fixed-step protocols within paired runs. \(^\ast\) marks \(p<0.001\) and ns marks \(p=0.42\) (two-sided paired Wilcoxon, 60 pairs per setting, unadjusted). The epoch\(^\dagger\) protocol runs \(10.5\times\) as many training steps, for reference only, and has no rank (--).}
    \label{tab:main}
    \setlength{\tabcolsep}{3pt}
    \begin{tabular*}{\dimexpr\linewidth-6mm\relax}{@{\extracolsep{\fill}}lccccc@{}}
        \toprule
        Setting & uniform & equal & new-only & SSCAL & \textcolor{gray}{epoch\(^\dagger\)} \\
        \midrule
        BSD10K / CLAP & \aulcresult{57.79}{0.67}{2.08} & \aulcresult{56.92}{0.53}{2.90} & \aulcresult{51.20}{0.18}{4.00} & \textbf{\aulcresult{59.79}{0.53}{1.02}}\(^{\ast}\) & \textcolor{gray}{\aulcresult{59.01}{0.42}{--}} \\
        BSD10K / PANNs & \aulcresult{50.70}{0.55}{1.65} & \aulcresult{49.66}{0.39}{2.77} & \aulcresult{39.40}{0.36}{4.00} & \textbf{\aulcresult{50.86}{0.62}{1.58}}\(^{\mathrm{ns}}\) & \textcolor{gray}{\aulcresult{52.21}{0.65}{--}} \\
        FSD50K / CLAP & \aulcresult{44.40}{0.60}{2.27} & \aulcresult{43.94}{0.55}{2.73} & \aulcresult{36.15}{0.60}{4.00} & \textbf{\aulcresult{48.95}{0.58}{1.00}}\(^{\ast}\) & \textcolor{gray}{\aulcresult{56.16}{0.33}{--}} \\
        FSD50K / PANNs & \aulcresult{45.69}{0.21}{2.01} & \aulcresult{45.42}{0.26}{2.88} & \aulcresult{36.43}{0.38}{4.00} & \textbf{\aulcresult{47.15}{0.17}{1.12}}\(^{\ast}\) & \textcolor{gray}{\aulcresult{43.46}{0.15}{--}} \\
        \midrule
        First / 240 & 34 & 4 & 0 & 202 & \textcolor{gray}{--} \\
        \bottomrule
    \end{tabular*}
\end{table*}

%% file: tables/t2_strategy.tex
\begin{table*}[t]
    \centering
    \small
    \caption{AULC (\(\%\)), mean\(\pm\)s.d. over ten seeds, each averaging four settings. The overall row also averages all six acquisition functions within each seed. Values combine BSD10K macro accuracy and FSD50K macro mAP. Parentheses give mean within-run ranks. The epoch\(^\dagger\) protocol runs \(10.5\times\) as many training steps, for reference only, and has no rank (--).}
    \label{tab:strategy}
    \setlength{\tabcolsep}{3pt}
    \begin{tabular*}{\dimexpr\linewidth-6mm\relax}{@{\extracolsep{\fill}}lccccc@{}}
        \toprule
        Acquisition & uniform & equal & new-only & SSCAL & \textcolor{gray}{epoch\(^\dagger\)} \\
        \midrule
        random & \aulcresult{51.44}{0.48}{1.98} & \aulcresult{50.84}{0.41}{2.90} & \aulcresult{44.64}{0.34}{4.00} & \textbf{\aulcresult{52.64}{0.50}{1.13}} & \textcolor{gray}{\aulcresult{53.00}{0.48}{--}} \\
        entropy & \aulcresult{45.28}{0.47}{2.08} & \aulcresult{44.77}{0.55}{2.68} & \aulcresult{33.17}{0.39}{4.00} & \textbf{\aulcresult{48.68}{0.65}{1.25}} & \textcolor{gray}{\aulcresult{50.32}{0.39}{--}} \\
        margin & \aulcresult{46.20}{0.50}{2.15} & \aulcresult{45.77}{0.51}{2.65} & \aulcresult{35.65}{0.80}{4.00} & \textbf{\aulcresult{49.64}{0.63}{1.20}} & \textcolor{gray}{\aulcresult{51.35}{0.63}{--}} \\
        Core-set & \aulcresult{49.86}{0.39}{1.78} & \aulcresult{48.60}{0.33}{2.98} & \aulcresult{40.07}{0.28}{4.00} & \textbf{\aulcresult{51.23}{0.53}{1.25}} & \textcolor{gray}{\aulcresult{52.36}{0.43}{--}} \\
        TypiClust & \aulcresult{53.22}{0.33}{1.85} & \aulcresult{52.52}{0.37}{3.00} & \aulcresult{46.00}{0.24}{4.00} & \textbf{\aulcresult{54.35}{0.33}{1.15}} & \textcolor{gray}{\aulcresult{54.63}{0.33}{--}} \\
        BADGE & \aulcresult{51.86}{0.46}{2.19} & \aulcresult{51.42}{0.24}{2.71} & \aulcresult{45.23}{0.24}{4.00} & \textbf{\aulcresult{53.58}{0.29}{1.10}} & \textcolor{gray}{\aulcresult{54.60}{0.41}{--}} \\
        \midrule
        overall & \aulcresult{49.64}{0.27}{2.00} & \aulcresult{48.99}{0.22}{2.82} & \aulcresult{40.79}{0.26}{4.00} & \textbf{\aulcresult{51.69}{0.32}{1.18}} & \textcolor{gray}{\aulcresult{52.71}{0.25}{--}} \\
        \bottomrule
    \end{tabular*}
\end{table*}

%% file: tables/t3_ablation.tex
\begin{table}[t]
    \centering
    \small
    \caption{Paired \(\Delta\)AULC versus uniform (points), mean\(\pm\)s.d. across ten seeds after averaging all 24 configurations within each seed. Ranks in parentheses compare seed means against uniform, equal and new-only within configurations (Tables~\ref{tab:main}--\ref{tab:strategy} use run ranks). Bold marks the best validation AULC.}
    \label{tab:ablation}
    \setlength{\tabcolsep}{2pt}
    \begin{tabular*}{\dimexpr\linewidth-6mm\relax}{@{\extracolsep{\fill}}lccc@{}}
        \toprule
        \(\rho\) & \(\beta=0.1\) & \(\beta=0.2\) & \(\beta=0.4\) \\
        \midrule
        0.6 & \aulcresult{+0.15}{0.11}{1.25} & \aulcresult{+0.16}{0.11}{1.38} & \aulcresult{+0.12}{0.14}{1.38} \\
        0.8 & \aulcresult{+0.48}{0.09}{1.08} & \aulcresult{+0.61}{0.08}{1.17} & \aulcresult{+0.70}{0.11}{1.21} \\
        1.0 & \aulcresult{+1.01}{0.12}{1.13} & \aulcresult{+1.50}{0.15}{1.21} & \textbf{\aulcresult{+2.04}{0.15}{1.13}} \\
        \bottomrule
    \end{tabular*}
\end{table}

%% file: tables/t4_diagnostics.tex
\begin{table}[t]
    \centering
    \small
    \caption{Default SSCAL, with B and F for BSD10K and FSD50K. Prec. is pooled positive-label precision and Cov. is \(|P_t|/|U_t|\), both in \(\%\), averaged with \(|P_t|\) over AL rounds and runs. \(\Delta\)Wall is training-side wall time (training + policy) over initialization and 40 AL rounds in excess of uniform (0.95--2.27\,s), averaged over 60 runs. SSCAL and uniform run 820 steps, epoch\(^\dagger\) 8,610 (\(10.5\times\), reference only).}
    \label{tab:diagnostics}
    \setlength{\tabcolsep}{2pt}
    \begin{tabular*}{\dimexpr\linewidth-6mm\relax}{@{\extracolsep{\fill}}lrrrrr@{}}
        \toprule
        Setting & Prec. & Cov. & \(|P_t|\) & \multicolumn{2}{c}{\(\Delta\)Wall (s)} \\
        \cmidrule(lr){5-6}
         & & & & SSCAL & \textcolor{gray}{epoch\(^\dagger\)} \\
        \midrule
        B / CLAP & 83.2 & 52.8 & 4092 & +1.36 & \textcolor{gray}{+7.43} \\
        B / PANNs & 75.4 & 48.0 & 3730 & +0.94 & \textcolor{gray}{+7.91} \\
        F / CLAP & 95.2 & 85.7 & 28035 & +1.36 & \textcolor{gray}{+9.85} \\
        F / PANNs & 87.2 & 71.3 & 23331 & +0.25 & \textcolor{gray}{+15.95} \\
        \bottomrule
    \end{tabular*}
\end{table}